Title

Internalized Morphogenesis: A Self-Organizing Model for Growth, Replication, and Regeneration via Local Token Exchange in Modular Systems

Authors

Takeshi Ishida[1*]

Affiliations

[1]Department of Ocean Mechanical Engineering, National Fisheries University
Shimonoseki, 759-6595, Yamaguchi, Japan.

* Takeshi Ishida
Address correspondence to: ishida@fish-u.ac.jp

Abstract

This study presents an internalized morphogenesis model for autonomous systems, such as swarm robotics and micro-nanomachines, that eliminates the need for external spatial computation. Traditional self-organizing models often require calculations across the entire coordinate space, including empty areas, which is impractical for resource-constrained physical modules. Our proposed model achieves complex morphogenesis through strictly local interactions between adjacent modules within the "body." By extending the "Ishida token model," modules exchange integer values using an RD-inspired discrete analogue without solving differential equations. The internal potential, derived from token accumulation and aging, guides autonomous growth, shrinkage, and replication. Simulations on a hexagonal grid demonstrated the emergence of limb-like extensions, self-division, and robust regeneration capabilities following structural amputation. A key feature is the use of the body boundary as a natural sink for information entropy (tokens) to maintain a dynamic equilibrium. These results indicate that sophisticated morphological behaviors can emerge from minimal, internal-only rules. This framework offers a computationally efficient and biologically plausible approach to developing self-repairing, adaptive, and autonomous hardware.

Keywords

Self-organization, Internalized Interaction, Token Model, Reaction-Diffusion, Swarm Robotics, Self-Replication, Regeneration, Turing Patterns, Cellular Automata

## 1. Introduction
### 1.1 Self-Organization and Engineering Applications

Self-organization, ubiquitously observed in biological systems, is a remarkable mechanism that generates dynamic and complex global order from limited local interactions. Applying this phenomenon in engineering to construct functional structures from numerous autonomous elements without central control is highly innovative for next-generation robotics, materials science, and nano/micro-engineering. The design methodology for "self-constructing" structures using such mechanisms has been systematized as Morphogenetic Engineering [1]. Recently, the field of "synthetic morphology" has also gained prominence, treating self-constructing architectures as design targets from the perspectives of synthetic biology and self-organized materials [2].

Implementing these concepts in swarm robotics or micro/nanomachines, which consist of a vast number of modules, imposes significant constraints. Since individual elements often possess extremely limited computational resources, communication bandwidth, and position-identification capabilities, a morphogenesis algorithm must satisfy two critical requirements: (i) operation via local communication only, and (ii) independence from calculations in the external coordinate space (i.e., empty areas outside the form). Consequently, achieving "self-organizing morphogenesis based solely on internal local interactions" is essential for the practical implementation of such systems.

**1.2 Turing (Reaction-Diffusion) Theory and the Current State of Morphogenesis Models**

One of the most influential mathematical frameworks for morphogenesis is Alan Turing's reaction-diffusion (RD) theory [3]. RD theory demonstrates that periodic structures, such as stripes and spots, emerge autonomously through the interactions of a few diffusible components, providing a fundamental basis for pattern formation research in developmental biology, evolutionary studies, and synthetic biology [4,5].

However, several challenges have long been discussed regarding the extension of RD to "morphogenesis": (a) managing pattern transitions as tissue domains grow, (b) facilitating the emergence of self-replicating structures, and (c) establishing a framework for regenerative recovery following structural perturbations.

For instance, RD on growing domains has been studied in detail regarding its influence on pattern selection, robustness, and mode conversion [6,7]. Experimental observations of self-replicating spots in chemical RD systems have been reported [8], and the underlying branching structures and transition dynamics have been theoretically analyzed [9]. Furthermore, Gierer–Meinhardt-type models, which represent local activation–long-range inhibition systems, are established as classic RD-based models for biological patterns [10]. Substantial research, including numerical simulations of *Hydra* head regeneration and the biological systematization of regenerative phenomena, has also been conducted [11,12].

In summary, while RD (Turing) systems have expanded their scope from "pattern formation" to "growth, self-replicating division, and regeneration," engineering implementation remains hindered by bottlenecks such as the requirement for a global computational domain (including empty space), high computational costs, and the feasibility of implementation under local communication constraints.

**1.3 Discretized RD Models and Implementation Feasibility under Local Communication Constraints**

Direct numerical solutions for reaction-diffusion (RD) systems are generally computationally intensive; therefore, lightweight discretization is preferable for implementation in swarm robotics and micro-nanomachines. One established approach involves mapping RD interactions onto Cellular Automata (CA). Young [13] formulated a CA model based on local activation and long-range inhibition to simulate vertebrate skin patterns. The author previously developed a totalistic CA model (the "Ishida model") that enables the creation and control of diverse patterns using specific morphological parameters [14].

However, most CA models, including the Ishida model, encounter two major challenges, namely communication costs and computational domains. First, transition rules typically require individual state counts of neighbors within a specific radius. This is a significant hurdle for simple modules that lack individual IDs and distance-based signal filtering. Second, calculations are typically conducted over the entire spatial domain. Most CA models update all cells, including empty "void" areas. In nature, biological morphogenesis occurs through internal interactions without computing the external space. Applying such algorithms to physical robot swarms would require unnatural external reference frames or waste resources in empty spaces. Furthermore, restricted internal computation often causes pattern stagnation at boundaries, hindering the development of complex appendages or segmented structures.

To address these issues, the author proposed the "Ishida token model" [15], which induces Turing instability solely through the exchange of integer values (tokens) and their temporal decay (aging). This model simulates RD-like dynamics by distributing tokens among adjacent modules, representing discrete approximations of substance concentrations, without solving partial differential equations. This allows pattern formation via communication only between directly adjacent robots, making it suitable for low-power microcontrollers.

Subsequent extensions using hierarchical local parameters have shown potential for complex structures [16]. Nevertheless, these models still rely on token propagation into external space or domain-wide state updates, leaving the second challenge unresolved. In the field of artificial life, continuous CA, such as Lenia [17] and Swarm Chemistry [18], demonstrate diverse self-organizing forms. While these are significant in showing diversity from local rules, a gap remains between their reliance on whole-space updates and the engineering requirement for strictly internal, local communication.

## 1.4 Research Trends in Self-Organizing Morphogenesis Based on Strictly Internal Local Interactions

Several approaches in artificial life and swarm robotics have explored self-organizing morphogenesis driven only by internal local interactions. A representative method involves implementing reaction-diffusion (RD) models directly on robots and exchanging morphogen concentrations through communication. Using the RD model, Slavkov et al. [19] reported a swarm robot control that achieves pattern formation without global coordinates through adjacent communication; however, clear morphological structures have not been fully realized. A previous study [20] applied the "Ishida token model" [15] to virtual modules; however, simply restricting the original model, which relied on external token exchange, to internal interactions made it difficult to maintain stable shapes or to generate complex segments and appendages.

Genetic Regulatory Network (GRN) models [21] emulate biological differentiation processes but tend to produce extremely complex rule sets. Digital hormone models [22] are effective for gradient-driven movement but struggle to maintain stable spatial patterns. Among the physical implementations, Rubenstein et al. [23] demonstrated shape formation using 1,000 Kilobots; however, this is essentially the "positioning" of modules based on a pre-programmed relative coordinate system. Amorphous computing [24] provides a theoretical foundation for pattern formation via coordinate-free local communication, although its characteristic is primarily that of a general computational theory.

In biology, several "internal-driven" models bypass external spatial computations. Vertex models [25] simulate epithelial segmentation and tubulation through the energy minimization of cell boundary vertices. Center-based models [26] treat cells as points, evolving colonies through short-range potentials and rules for division and death. The subcellular element method [27] represents single cells as particle clusters, confining the calculations to cell interiors and boundaries. Growth tensor models [28] perform computations exclusively within a material mesh, expressing wrinkling and budding through elastic relaxation. While these models accurately reflect biological insights, their direct implementation in engineering micromachines is difficult, as they require individual elements to perform sophisticated physical calculations or global energy minimization.

**1.5 Research Objectives and Novelty of This Study**

Based on the research discussed above, the challenges in this field can be summarized as follows:
(1) While knowledge regarding reaction-diffusion (RD) and Turing systems is extensive (including RD on growing domains, self-replicating spots, and regeneration models), the requirements for whole-space computation and high computational loads remain significant obstacles to implementation in engineering modular swarms.
(2) Although Cellular Automata (CA) and discrete models possess high suitability for implementation, many existing designs depend on updating the external space where no physical entities exist or require distance-dependent information filtering, making them difficult to apply to realistic module groups.

To address these issues, this study builds upon the token-exchange CA (the Ishida token model) to present a minimal model for "internalized morphogenesis." This model is driven solely by "adjacent interactions within the body" (the region where modules exist) and eliminates computation in external space. The proposed framework handles the growth, division (equivalent to self-replication), and post-amputation recovery (equivalent to regeneration) within a single unified scheme. It should be noted that this study does not claim to be the "first" in morphogenesis research applying RD (Turing) theory. Rather, while respecting the historical progress of the field (e.g., research on growth domains, self-replicating spots, and regeneration) [6–12], this study focuses on presenting these phenomena as a concise computational framework under the specific constraints of engineering implementation: internal-only processing, local communication, and the absence of external spatial computation.

Specifically, the goal is to establish a model for engineering applications that expands the previously developed token-exchange model (Ishida token model) [15] to "internally restricted actions." This enables the growth, replication, and regeneration of morphology through adjacent interactions and individual module parameters alone, without referencing any information from the external space. The "token-exchange model"[15] adopted in this study controls morphology through the transfer of discrete values (tokens). The token quantity functions as a discretized approximation of "substance concentration" in continuous models. Specifically, by distributing tokens among adjacent modules according to fixed rules (equivalent to diffusion coefficients), dynamic pattern formation similar to reaction-diffusion systems is realized without directly solving partial differential equations. This results in a low-load computational model that is executable, even on inexpensive microcontrollers.

The novelty of this study is concentrated in the following points:

First, the model utilizes only the "internal propagation" of tokens to the adjacent modules. Tokens are exchanged only within the region where the modules are present, and the "extension (growth)" or "contraction (atrophy)" of the boundary modules is determined based on the potential values derived from that distribution. Although previous research [20] applied this model to virtual modular robots, it faced difficulties in maintaining stable shapes or forming complex segments and appendages owing to unstable potential values. The present study achieved advanced morphogenetic capabilities by revising the token exchange algorithm and incorporating growth and contraction thresholds at boundaries.

Second, the model offers simplicity that does not require sophisticated communication functions. Although advanced micromachines emulating hormone transmitters are being conceptualized, early-stage microrobots can only be expected to share limited information, such as the exchange of numerical values. This model can realize pattern formation via Turing instability and the subsequent morphological extension, self-division, and regeneration through simple numerical interactions alone.

Furthermore, this model does not depend on spatial metrics; it represents spatial patterns using only topological adjacency and temporal decay (aging). This research presents a true "self-organizing morphogenesis model" that does not rely on external spatial computation, thereby opening possibilities for constructing complex structures in systems such as micro- and nano-robot swarms where communication and computational resources are extremely limited.

## 2. Model

The morphogenesis model proposed in this study was designed to autonomously transform morphology based on the internal potential value distributions. These potential values were calculated from the local token exchange rules between individual modules and the number of tokens recorded within each module. This section provides a detailed description of the spatial definition of the model, the token propagation algorithm, the potential value calculation method, and algorithms governing morphological changes.

### 2.1 Computational Space

The computational space in this model is defined as a hexagonal lattice consisting of 150 × 150 cells (or an arbitrary size depending on the simulation settings) (Fig. 1A). Each cell (i, j) assumes one of the following two states. Periodic boundary conditions are applied at the boundaries of the computational space. Hereafter, each cell whose state is 1 is referred to as a "module."

**State 0:** "External space (empty region)" where no module exists.
**State 1:** "Internal region (body domain)" where a module exists.

A hexagonal lattice was employed because the distances between adjacent cells are more uniform than in a square lattice, enabling a more faithful representation of biological cell packing and pattern formation with minimal anisotropy. The coordinates of each cell are horizontally offset between odd and even rows, establishing uniform neighborhood relationships in all directions.

In the initial state of the simulation, state 1 cells were arranged in a circular pattern with radius R at the center of the space, whereas the remaining regions are set to state 0 (Fig. 1B). The most significant feature of this model is that the morphology is determined solely by operations within the state 1 cells (internal region); specifically, no transition rule evaluations or information updates are performed in the state 0 regions. Consequently, this model demonstrates that morphology emerges exclusively from internal interactions within the individual, independent of the external environment.

**2.2 Inter-module Interaction (Token Propagation Algorithm)**

Information transfer between modules is implemented using a "token model," which emulates reaction–diffusion dynamics through the exchange of discrete integer values (tokens). A schematic representation of the model is shown in Fig. 1C. In this study, an algorithm is employed where information remains strictly confined within the morphology by distributing tokens among modules within the body domain.

**2.2.1 Token Generation and Propagation Rules**
**Basic Model**

In each time step of the simulation, all modules currently in State 1 execute a computational cycle consisting of the following five stages (1 – 5):

1. **Token Generation**: Each State 1 module generates a single token with an initial token amount of 1 and an aging label (elapsed steps) of $v = 1$.
2. **Sequential Propagation**: The generated tokens are distributed equally among the six adjacent modules at each step. This propagation occurs sequentially for up to Z iterations (where Z is a parameter defining the maximum propagation distance; this study set $Z = 20$).
3. **Token Classification**: Two types of tokens are defined during the propagation process:

   (a) **Propagating Tokens**: These tokens represent the "wavefront" of substance diffusion and move between modules. In each exchange, the label v of the moving token increments (e.g., from 2 to 3). Conversely, since the token amount is divided during each distribution, the concentration of tokens with higher labels becomes extremely low, which effectively functions as an elimination process.

   (b) **Accumulated Tokens**: Each module adds the amount of "propagating tokens" received from adjacent modules to its own "accumulated tokens" for each label v. This accumulated quantity serves as the basis for the potential calculations. These tokens are not reset during the Z iterations. Each module also maintains its own self-generated tokens as a part of this accumulation.
4. **Distribution and Boundary Conditions**: The amount of tokens with label v is distributed equally (i.e., 1/6 of the current amount) to cells in the six neighboring directions. A token is transferred only if the adjacent cell is in State 1 (internal region), at which point the label is updated to $v+1$ for further accumulation and re-propagation. If an adjacent cell is in State 0 (external space), the tokens distributed in that direction are assumed to vanish. This "outflow from the boundary" automatically forms a token concentration gradient corresponding to the geometric shape of the morphology.

5. **Potential Calculation and Morphological Update**: Following the Z propagation iterations, the potential value P of each module is calculated based on the accumulated token volume, as described in Section 2.2.2. Based on these values, the growth and contraction processes detailed in Section 2.3 are applied to the boundary modules.

**Time-Series Inheritance Model**

While the basic model resets the token distribution at the start of each time step, an inheritance model is also introduced to preserve the distribution from the previous step to account for biological continuity. In this model, at the start of the next step, the accumulated token volume in all cells is multiplied by the transfer rate k ($0 \leq k \leq 1$) and added to the newly generated tokens with v = 1 before the propagation cycle begins. Tokens distributed to State 0 are not carried over (i.e., they are discarded upon cell disappearance). This mechanism allows past morphological information to influence the current dynamics through temporal filtering.

The rationale for this model is the anticipated difficulty of resetting information at every time step when implementing this framework in micro-scale systems in the future. This approach is expected to be more readily transferable to micro- or molecular systems.

**2.2.2 Calculation of Potential Values**

The internal potential *P(i, j)* for each module (i, j) was calculated as a weighted sum based on the aging label v of the accumulated tokens. Specifically, the total token volumes within the following two ranges were compared:

*sumX*: the sum of accumulated tokens with aging labels *v* from 1 to *X*.

*sumY*: the sum of accumulated tokens with aging labels *v* from 1 to *Y*.

where *X < Y*. Given these values, the potential *P* is derived using the morphological determination parameter *w* as follows:

$$P = sumX - (sumY \times w) \qquad (1)$$

This equation is an extension of the state transition rule used in the Ishida model, where decisions are based on the value of (sum of states in the inner neighborhood) - (sum of states in the outer neighborhood) × *w*. In this context, *sumX* and *sumY* correspond to inner and outer neighborhoods, respectively. This formula essentially maps the "local activation–long-range inhibition" mechanism of the reaction-diffusion theory from the spatial domain to the temporal domain (aging) via token residence time, thereby achieving a pseudo-representation of such dynamics. By adjusting the value of *w*, various potential gradients—such as spots, stripes, or complex hierarchical structures—can be generated within the individual morphology.

**2.3 Dynamics of Morphological Transformation**

Each module autonomously executes decision rules for "growth (state 0 → 1)" and "degeneration (state 1 → 0)" based on the spatial distribution of the calculated potential values P (Fig. 1C). In the proposed model, the degeneration rule (state 1 → 0) is applied universally to all modules within the internal domain (state 1), rather than being limited to boundary modules. Conversely, regarding expansion via the growth rule (state 0 → 1), two distinct configurations are considered: one where it is restricted to boundary modules in contact with the external space, and another where it is also applied to the internal domain.

### 2.3.1 Degeneration Algorithm (State 1 → 0)

Each module within the internal State 1 domain calculates the potential $P$ at each time step. To maintain cell viability, P must remain within the appropriate range (survival domain). A module undergoes "death" and transitions to State 0 if either of the following conditions is satisfied:

- Condition A (lower survival limit): $P \leq R$  (when the potential is too low, indicating insufficient activation).

- Condition B (upper overcrowding limit): $P \geq R_o$  (when potential is too high, indicating an overcrowded or saturated state).

In this context, $R$ is the survival threshold below which a module dies, and $R_o$ is the threshold representing saturation or overcrowding. Under these rules, regions within the internal structure corresponding to the inhibitory zones were removed as physical voids. This process triggers complex morphological transformations, such as the formation of tubular structures and self-division phenomena.

### 2.3.2 Growth Algorithm (State 0 → 1)

The morphological expansion is driven by boundary modules (State 1 modules that have at least one State 0 neighbor). When the potential value P of a boundary module exceeds the growth threshold $G$ ($P \geq G$), the cell generates (extends) a new module in the surrounding empty space.

1. Selection of extension candidates: One neighboring cell is randomly selected from adjacent State 0 cells.
2. Conflict resolution: If multiple boundary modules attempt to extend into the same empty cell simultaneously, one boundary module is prioritized at random, and the target cell is updated to state 1.

Through this process, the increase in internal potential values manifests as the physical extension of "limbs" or general "growth" of the morphology. As growth and degeneration proceed in parallel, the form evolves toward a dynamic equilibrium or specific geometric configurations.

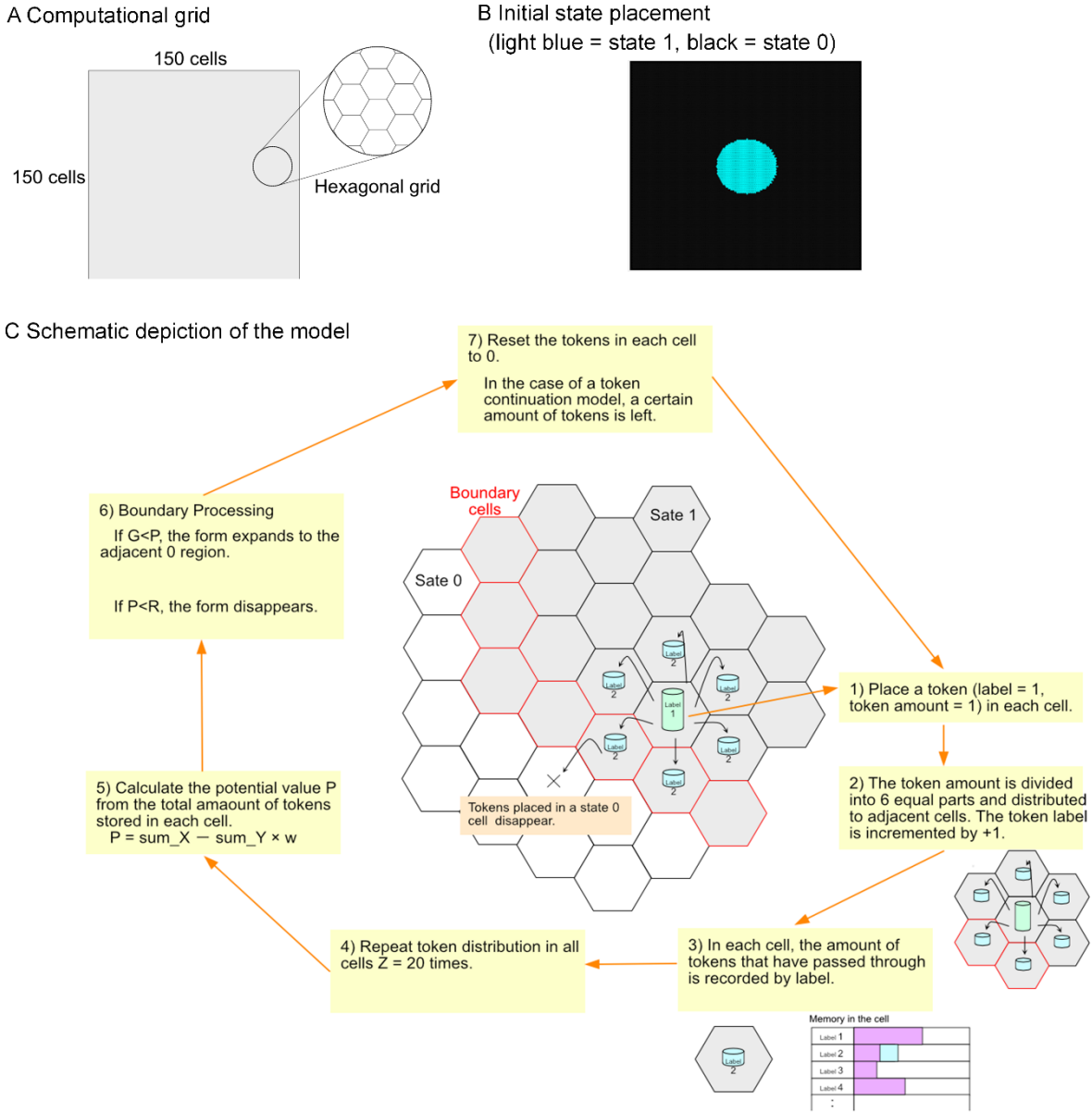

**Fig. 1 Schematic diagram of the model and initial state placement.** **A** The computational grid is a hexagonal lattice consisting of 150 × 150 cells. **B** Initial state placement: cells within a 20-cell radius circle at the center of the computational domain are set to State 1. **C** Flow of computational processes at each time step.

**2.4 Model Implementation and Simulation Environment**

The proposed model was implemented using JavaScript and HTML5, with a user interface designed to allow the configuration of various parameters. This setup enables the real-time computation of individual conditions. The pseudocode for the model is provided in Appendix 9, and the actual source code can be accessed via the URL specified in the DATA AVAILABILITY section.

Regarding random numbers, a pseudo-random number generator (JavaScript Math.random() with a fixed seed)

was used to select growth candidates and resolve conflicts. This approach facilitates symmetry breaking from the initial state and promotes anisotropic growth.

## 2.5 Quantitative Evaluation Indices

To quantitatively analyze the morphogenesis process, the following evaluation metrics were defined and recorded as time-series data:

1. Area (Active Cell Count): The total number of cells constituting the individual. This is used to evaluate the growth rates and the final size.
2. Perimeter (Boundary Cell Count): The total number of boundary cells in contact with external space.
3. Circularity: An index calculated as $4 \times \pi \times$ Area / Perimeter$^2$. Values closer to 1 indicate a circular (stable) form, whereas lower values represent complex branching or extension.
4. Dispersion: The average distance from the centroid of the morphology to each constituent cell. This evaluates the spread of the form and the breakdown of symmetry.
5. Potential distribution characteristics: The maximum, minimum, and average values of potential $P$ within the individual.

By utilizing these indices, the transition process from the initial state to the steady state, as well as the influence of morphological parameters such as $w$ on morphological diversity, can be objectively evaluated.

## 3. Results
## 3.1 Determination of Reference Parameters and Basic Morphogenesis

The proposed model involves numerous parameters, including the morphological parameter $w$, growth threshold $G$, survival threshold $R$, and ranges $X$ and $Y$, resulting in an immense combinatorial space. However, based on the findings from previous CA models, significant morphologies are expected to emerge in only a small fraction of these combinations. Therefore, the parameter regions yielding meaningful morphologies have been identified through several global trials. A single reference parameter set was determined, as presented in Table 1 (with the growth decision process restricted to boundary modules as the baseline).

Under this reference parameter set, the initial circular configuration immediately shrank upon starting the simulation, leading to a branched state where "limbs" extend (Fig. 2A). Regarding the distribution of potential values (Fig. 2B), the tips (growth points) exhibited high potential (red), whereas the stems and internal regions remained stable with intermediate potential (yellow/green), and the boundaries showed low values (green/blue).

The time-series results for each index are presented in Supplementary Figure A1 (Appendix 1). As the limbs extended, both the area and perimeter increased, circularity significantly decreased initially, and dispersion gradually increased over time.

**Table 1 Reference Parameters**

| Z (Token Steps) | 20 |
|---|---|

| X | 8 |
|---|---|
| Y | 16 |
| w | 0.470 |
| G (Growth Threshold) | 1.40 |
| R (Survival Threshold) | 1.27 |
| Ro (Overcrowding Threshold) | 50 |

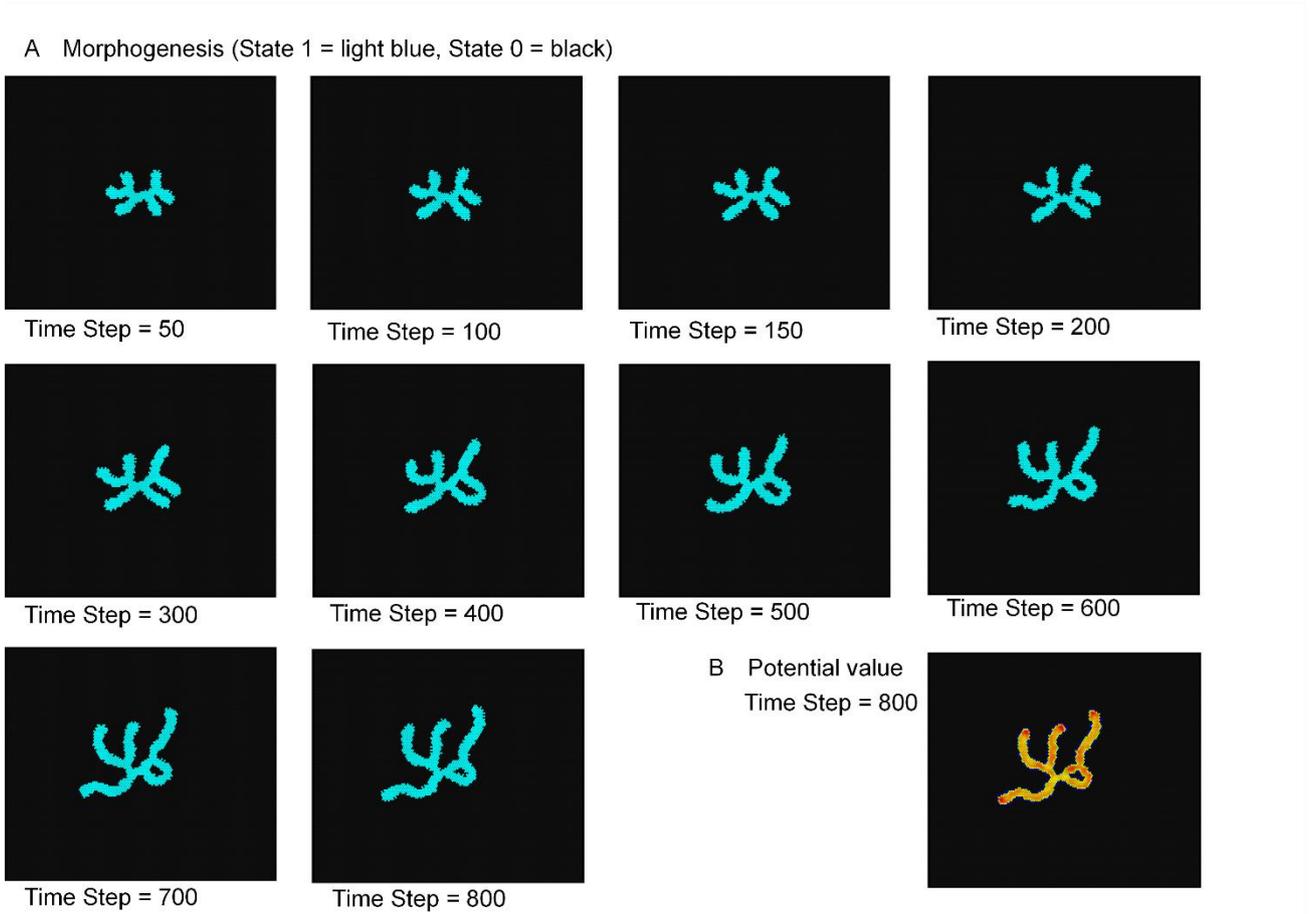

**Fig. 2 Morphogenesis results under the reference parameter set (time-series) and potential value distribution at 800 time steps.** **A** Pattern showing the slow extension of limbs. **B** Distribution of potential values: tips (growth points) exhibit high potential (red), stems and internal regions show stable intermediate potential (yellow/green), and boundaries show low values (green/blue).

### 3.2 Parameter Sensitivity and Phase Diagrams
#### 3.2.1 Influence of the *w* Parameter

Table 2 shows the results of varying the *w* parameter relative to the reference parameter set (at 400 time steps). Compared to the reference value (*w* = 0.470), increasing *w* resulted in weaker limb extension. For *w* ≥ 0.480, the initial region shrank. Although there is a tendency toward extension, the structure continues to contract and persists only as a small cluster. Conversely, when *w* is smaller than the reference value, the boundary tends to expand. At *w* = 0.450, black spots formed incidentally when the morphology collided with the periodic boundary, replicated internally and spreading throughout the domain (Supplementary Appendix 2, Figure A3). When $w \leq 0.430$, the boundaries expanded to the point of collision, and any incidentally formed black spots failed to grow.

Table 2 Index values when the *w* value is changed for the reference parameter set (results at 400 steps)

| w | 0.44 | 0.45 | 0.46 | 0.47 | 0.48 |
|---|---|---|---|---|---|
|  | 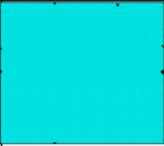 | 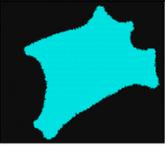 | 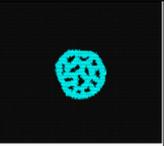 | 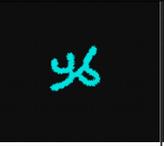 | 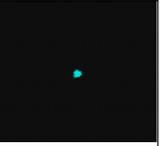 |
| activeCellCount | 22465 | 8714 | 1597 | 894 | 46 |
| boundaryCellCount | 56 | 518 | 517 | 353 | 24 |
| potentialAvg | 0.97 | 0.95 | 1.54 | 1.55 | 1.62 |
| potentialMax | 1.65 | 2.07 | 1.94 | 2.07 | 2.03 |
| potentialMin | 0.96 | 0.72 | 0.71 | 0.71 | 0.71 |
| circularity | 90.02 | 0.41 | 0.08 | 0.09 | 1.00 |
| dispersion | 268.0 | 176.9 | 79.8 | 75.2 | 12.1 |
| activeCell/boundaryCell | 401.16 | 16.82 | 3.09 | 2.53 | 1.92 |

### 3.2.2 Phase Diagram of Growth Threshold *G* and Survival Threshold *R*

Figure 3 presents the phase diagram at 400 time steps for varying values of growth threshold *G* and survival threshold *R*.

When *G* is varied while keeping *R* at its reference value, *G* = 1.30 and *G* = 1.35 result in limb extension that eventually leads to fusion between limbs and the formation of internal black spots. At the reference value of *G* = 1.40, a limb-extending pattern is observed. At *G* = 1.45, the morphology persisted in a stationary state without extension. When *G* ranges from 1.50 to 2.0, the initial region shrinks into a small cluster and forms a fixed pattern.

When *R* is varied while keeping *G* at its reference value, *R* = 1.07 yields a fixed pattern close to the initial circular configuration; decreasing *R* further causes the morphology to remain almost perfectly circular. At *R* = 1.12, limb extension ceased mid-growth. For *R* = 1.17 and *R* = 1.22, limbs extend but stop growing upon collision with one another. Notably, growth was observed to be most active at *R* = 1.17. At the reference value of *R* = 1.27, the limbs extended gradually, whereas for *R* = 1.32 and *R* = 1.37, they grew very slowly and incrementally.

Overall, small values of *G* and *R* lead to the expansion of mesh-like patterns across the entire space, whereas large values result in small, fixed circular patterns. Patterns with limb extensions emerged in the intermediate region between these extremes. Comprehensive investigations of other parameters are provided in Appendices 3 through 6 of the Supplementary Materials.

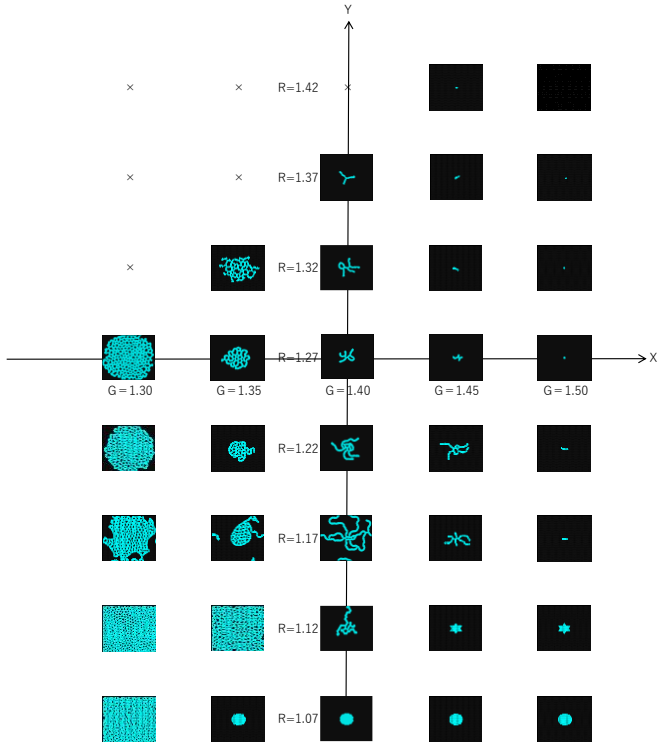

**Fig. 3 Phase diagram of results for varying the growth threshold *G* and the survival threshold *R* relative to the reference parameter set (at 400 time steps).**

### 3.3 Self-Replication and Regeneration
#### 3.3.1 Search for and Results of Self-Replicating Parameters

A search was conducted to identify the parameter sets that induced the self-replicating patterns. Significant self-replication was observed under the following configuration: $w = 0.50$, $X = 9$, $G = 1.37$, $R = 1.35$, $Ro = 1.70$ (other parameters were identical to the reference set).

The time-series results for the self-replicating pattern are shown in Figure 4. In typical Turing pattern models, patterns exhibit a repulsive nature owing to the interaction between long-range inhibition and short-range activation, which prevents them from merging. However, because the proposed model relies exclusively on internal interactions, the extended morphologies do not repel each other and tend to fuse immediately upon contact. Under the specific parameter set mentioned above, a balance between growth and degeneration allows the structure to multiply and proliferate without fusion.

Varying *Ro* by increments of 0.01 significantly altered the outcomes. At $Ro = 1.71$, the mesh structure expanded and subsequently fragmented. Between $Ro = 1.72$ and $1.75$, a stable mesh pattern formed. Conversely, at $Ro = 1.69$, self-replication occurred occasionally, but most trials resulted in either the maintenance of a single cluster or total extinction. At $Ro = 1.68$, the system typically led to extinction or persistence of a minimal cluster.

In this model, the survival threshold *R* defines the limit below which a region vanishes, whereas the overcrowding threshold *Ro* defines the limit above which a module dies owing to saturation. In the reference parameter set, where *Ro* = 50, this threshold remained inactive. However, lower *Ro* values had a significant influence. This sensitivity to *Ro* mirrors the rules of Conway's Game of Life, where survival depends on a moderate range of activation. Similar to the "Ishida model," this mechanism enables the emergence of diverse patterns, including self-replication. A detailed investigation of self-replication using the "inheritance model," which retains a portion of tokens from the previous step, is provided in Appendix 7.

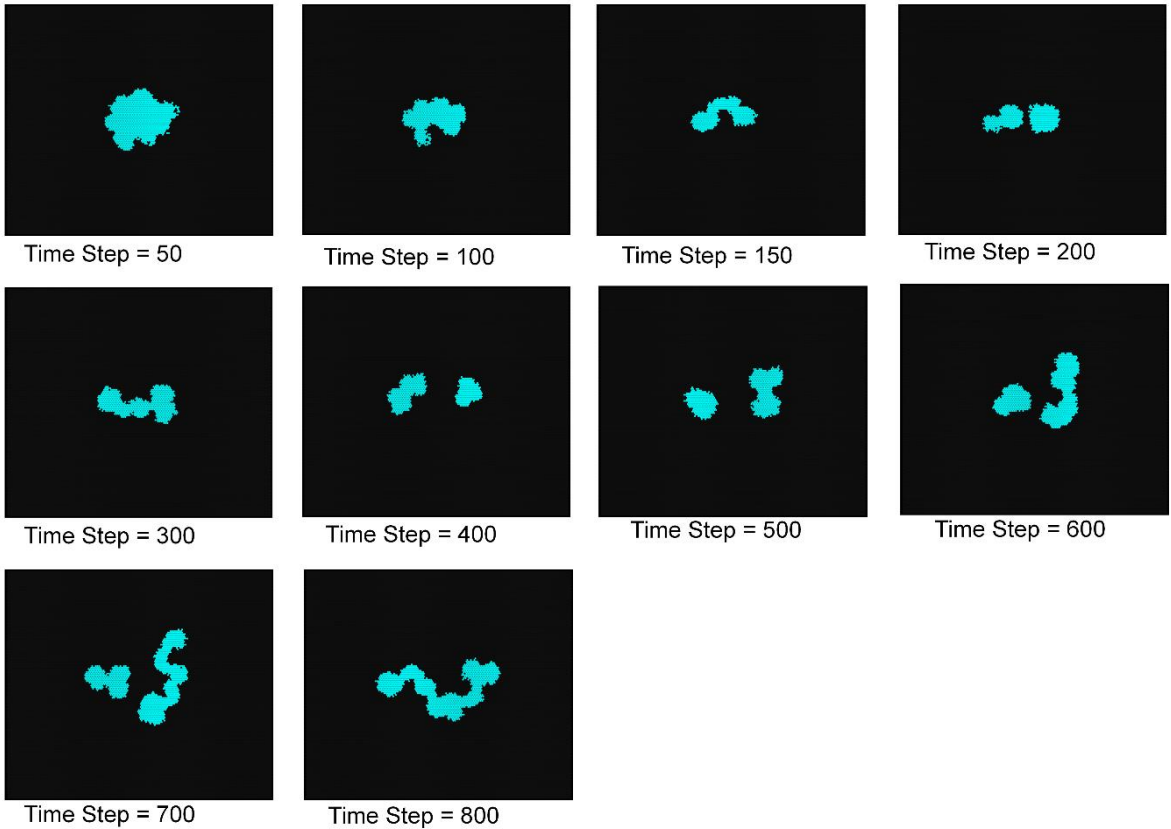

**Fig. 4 Time-series results of the self-replicating pattern; *w* = 0.50, *X* = 9, *G* = 1.37, *R* = 1.35, *Ro* = 1.70 (other parameters are identical to the reference parameter set).**

### 3.3.2 Investigation of Self-Regeneration Capabilities

An investigation was conducted to determine whether the proposed model could generate regenerative morphological patterns. Specifically, the simulation was programmed to set the upper half of the computational space (cells within the range of *y* = 1 to 65) to state 0 at time step 200, effectively amputating the morphology of state 1. Subsequently, the recovery of the form was observed. Since the limbs extend very slowly under the reference parameter set, specific parameters that induce a more rapid morphogenetic process were identified for this analysis.

Figure 5 shows the results using *G* = 1.42 and *R* = 1.2 (with other parameters remaining identical to the reference set), representing a pattern where limbs extend rapidly. Furthermore, Figure 6 illustrates the results for the parameters where the mesh-like morphology expanded (*G* = 1.30, *R* = 1.2). In the limb-extension pattern, the limbs were observed to re-extend following amputation. However, it was also noted that if the limbs fused to form a circular ring structure, the morphology occasionally became fixed, preventing further limb emergence. In the mesh-expansion pattern, it was confirmed that the morphology gradually returned to its original state after partial removal.

The time-series transitions of each evaluation index for the cases shown in Figures 5 and 6 are provided in Appendix 8. These results demonstrate that the area, perimeter, and dispersion recovered consistently following amputation at step 200. To quantify the degree of regeneration, a Recovery Index (*RI*) was defined as follows:

$$RI = \frac{V(T) - V_{post}}{V_{pre} - V_{post}}$$

where *Vpre* represents the dispersion value immediately before amputation, *Vpost* is the value immediately after amputation, and *V(T)* is the value after *T* steps from amputation. In this study, "successful regeneration" was defined as $RI \geq 0.8$ (80% recovery). In the limb amputation case (Figure 5), the *RI* reached 0.8 within 55 steps. In the mesh amputation case (Figure 6), the *RI* reached 0.8 within 36 steps.

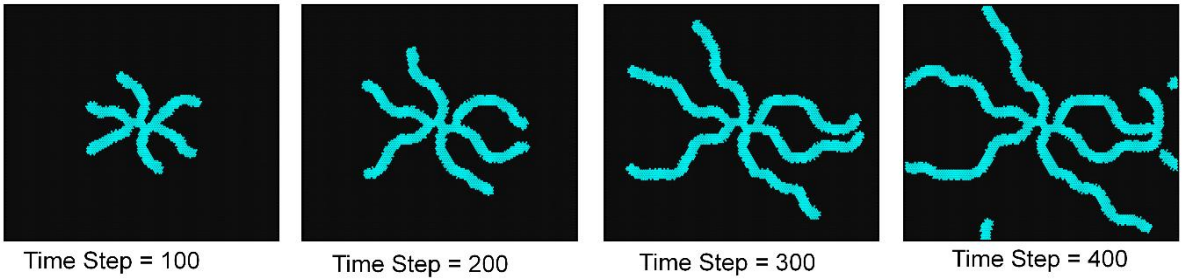

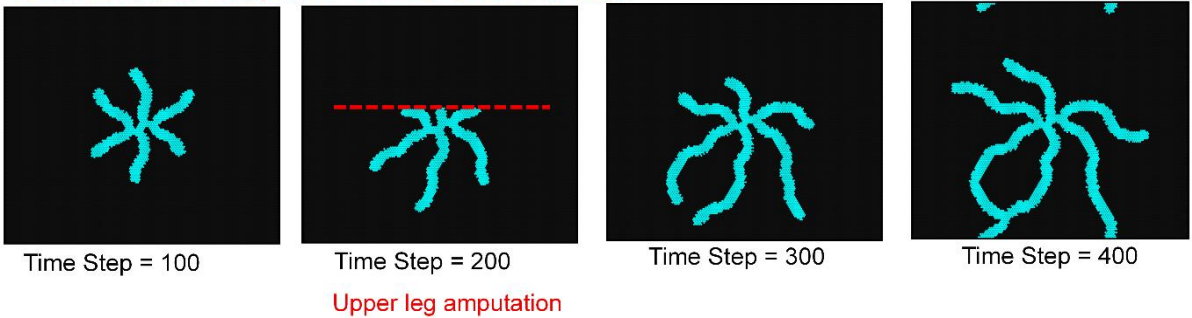

**Fig. 5 Time-series results of morphological regeneration patterns (limb extension, amputation, and regeneration).** (*G* = 1.42, *R* = 1.2; other parameters identical to the reference set). **A** Without amputation. **B** With amputation: upper half removed at 200 time steps.

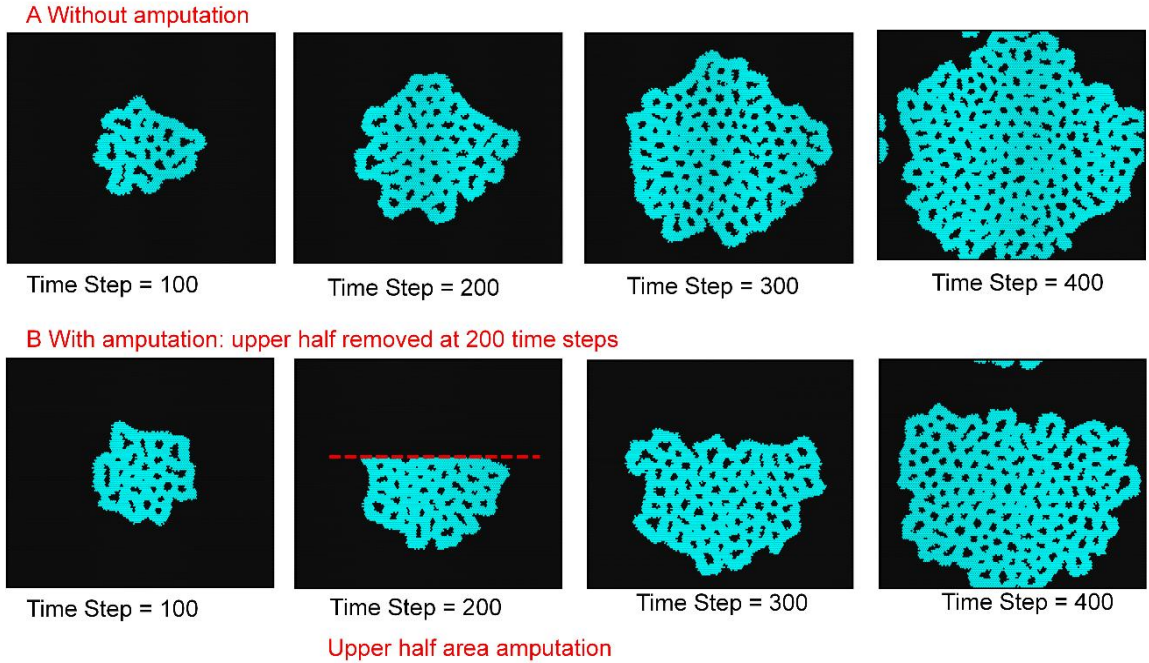

**Fig. 6 Time-series results of morphological regeneration patterns (mesh expansion, amputation, and regeneration).** ($G = 1.30$, $R = 1.2$; other parameters identical to the reference set). **A** Without amputation. **B** With amputation: upper half removed at 200 time steps.

## 4. Discussion
### 4.1 Discussion of the calculation results
#### 4.1.1 Comparison with Models for Generating Turing Patterns

In the Ishida model, variations in the morphological parameter $w$ determine the resulting configuration: at low $w$ values, the state 1 (blue) region expands and develops state 0 (black) spots; at intermediate $w$ values, stripe-like patterns are formed; and at high $w$ values, the system produces isolated state 1 (blue) spot patterns.

The proposed model exhibits similar tendencies as a function of $w$ (Table 2); at low $w$ values, the State 1 domain undergoes expansion. As illustrated in Figure A3 of Appendix 2, it has been observed that if State 0 spots incidentally emerge within the State 1 region, they undergo self-replication and subsequently proliferate throughout the domain.

Furthermore, the proposed model is characterized by the emergence of limb-like extensions. These limbs eventually collide with each other to form mesh-like architectures. This limb extension process is interpreted as a transitional phase toward the formation of a mesh structure, which corresponds to the emergence of stripe patterns in conventional Turing systems.

#### 4.1.2 Symmetry Breaking and the Role of Fluctuations

Despite an initially isotropic circular configuration, symmetry breaking occurred, characterized by the extension of "limbs" with distinct directionality over time. This phenomenon is attributed to the amplification of fluctuations during the expansion and shrinkage processes of modules, driven by boundary processing on a discrete hexagonal lattice through reaction-diffusion-like instabilities. This mechanism is considered to be applicable to pattern formation in both the Ishida model and the proposed framework. In biological development, it is well established that fine-scale fluctuations at the cellular level act as triggers for global polarity determination, and the present model demonstrates that such processes can be reproduced solely through intrinsic, internalized computation.

### 4.1.3 Analysis of Potential Value Distribution

The distribution of potential values under the reference parameters (Fig. 2B) indicates that the tips, which function as growth points, exhibit a high potential (red). In contrast, the stems and internal regions remain stable with intermediate potential (yellow/green), while the boundary areas showed low values (green/blue). Within the proposed algorithm, the potential value $P$ is determined by the equation $P = sumX - (sumY \times w)$. Since the tips project into the external State 0 space, they receive a minimal inflow of "old tokens" ($sumY$, representing the inhibitory factor) from the surrounding environment. Meanwhile, "new tokens" ($sumX$, representing the activating factor) generated by the modules themselves are maintained. Consequently, the inhibitory term ($sumY \times w$) remains small, resulting in a high potential $P$.

Conversely, the internal and stem regions were entirely surrounded by other modules. This configuration facilitates the accumulation of "old tokens" ($sumY$) propagated from neighboring cells, thereby increasing the inhibitory term ($sumY \times w$). This causes the potential $P$ to decrease below the growth threshold $G$, effectively preventing further structural thickening. This mechanism can be regarded as functionally equivalent to the process of generating Turing patterns driven by the interaction between activators and inhibitors. These observations suggest that even a model restricted to internal interactions can successfully produce states analogous to those found in reaction-diffusion systems.

### 4.1.4 Role of Boundary Conditions as a Field for "Information Dissipation"

A critical consideration in the proposed model is that, while the external space (State 0) is not utilized as a computational resource, it functions as a "sink" for the disposal of unnecessary tokens—effectively acting as a field for "information dissipation." The outflow of internally generated, aged tokens (information) through the boundaries ensures the maintenance of the potential gradients necessary for internal order formation. This mechanism is consistent with thermodynamic principles, in which living systems maintain a low-entropy state (order) by exporting entropy to their environment. Thus, the model explicitly demonstrates the physical significance of the "boundary" within self-organizing systems.

### 4.1.5 Parameter Sensitivity

The morphologies that emerge when varying the parameter $w$ and the growth threshold $G$ include ordered states such as stationary circular clusters, limb-like extensions, and morphological expansion throughout the space. In most other regions of the parameter space, the entire domain typically converges to a uniform state of either state 0 or state 1. This observation is consistent with the phenomenon in complex systems science where complex computation and life-like behaviors emerge exclusively at the "edge of chaos." This suggests that in the proposed model, dynamic morphogenesis is maintained only at critical points where the balance between activators (tokens with low label counts) and inhibitors (tokens with high label counts) is in equilibrium, avoiding both "static death" (disappearance of the form) and "dynamic death" (expansion of the form across the entire domain).

Furthermore, the evaluation metrics in Table 2 suggest that meaningful configurations can be identified through combinations of circularity, the area-to-perimeter ratio, and dispersion. Specifically, a form may be considered significant when the circularity is low (approximately 0.01), dispersion is at an intermediate level (approximately 70), and the ratio of active cells to boundary cells is between 2 and 3. These indices could enable the automated exploration of significant forms, such as through the use of genetic algorithms (GA) to search for parameters. As this report examines only a subset of the parameter space, a more extensive search may lead to the discovery of even more diverse morphologies. The results demonstrate that behaviors can shift drastically with a change of only 0.01 in *w* or *Ro*; this parameter sensitivity reflects the inherent phenotypic diversity of the system, suggesting high suitability for automated search methodologies using GA.

**4.1.6 Analysis of the Token Inheritance Model**

It was found that even a model that retains a fixed proportion of tokens rather than resetting them at each time step (results presented in Appendix 6) can emerge Turing-like morphologies in a manner similar to the basic model. This implies that the inheritance model is highly suitable for practical applications in micro-robotic or molecular systems. The rationale for this is that resetting information at every time interval is expected to be technically challenging when implementing a "general-purpose morphogenesis model based on internal interactions" at the micro-scale in the future; therefore, a model that allows for token persistence offers greater feasibility for such platforms.

**4.1.7 Investigation of Self-Replication and Self-Regeneration**

While the proposed model does not generate perfectly regular self-replicating patterns as seen in the Ishida model, it was possible to produce instances where domains fragment through parameter adjustment, although these domains occasionally fuse or separate. A future challenge involves implementing mechanisms to prevent boundary fusion, such as the introduction of a "skin" layer or similar constraints.

Furthermore, regarding morphological self-regeneration, it was demonstrated that limbs can regenerate provided that a protrusion remains after amputation. Regeneration of mesh-like regions was also confirmed. The behavior in which amputated limbs re-extend or recover their original mesh structure indicates that the model does not rely on a static "blueprint." Instead, it generates morphology in a dynamic "equilibrium state" (attractor). Even when part of the system is lost, the balance of token generation, propagation, and outflow in the remaining section is recalculated, leading to a reconfiguration of the potential field. This implies that the characteristic of biological systems as a "dynamic equilibrium"—which maintains order amongst the continuous turnover of matter, as noted by Schrödinger and Rudolf Clausius—has been successfully implemented in an engineering context using simple local rules.

**4.2 Scholarly Significance and Engineering Applications of the Proposed Model**
**4.2.1 Significance of Self-Organization Independent of External Computation**

Many conventional reaction-diffusion models and cellular automata (CA) require computation over the entire lattice, including empty external space (voids). In contrast, the proposed model completes the morphogenetic process solely through token communication and storage within "internal" modules. This implies that in swarm robotics or micro/nanomachines where computational and memory resources are extremely limited, self-organization can be realized without referencing information from regions where no robots exist, which significantly enhances the feasibility of physical implementation. The outflow of tokens to the external environment naturally generates potential gradients between the morphological boundaries and interior. This mechanism, which contributes to the control of curvature and protrusions, is highly consistent with the behavior of biological morphogens.

### 4.2.2 Emergence of Self-Replication and Self-Repair Capabilities

The self-replication phenomena demonstrated in the computational results (Section 3.3) suggest that individuals can be maintained and proliferated based solely on local potential distributions without central control. Furthermore, the process in which internal inhibitory potentials induce a "constriction" once the morphology exceeds a certain size, leading to physical fission, serves as an example of global order (reproduction) emerging from a single rule set. Moving forward, quantitative investigations of self-replication are necessary. Potential approaches include measuring the number of consecutive time steps during which separated connected components persist as independent individuals, or quantitatively evaluating the area of each individual following separation.

Additionally, the model successfully exhibited "self-regeneration (self-repair)" behavior; even when a portion of the morphology is artificially removed, tokens are redistributed among the remaining modules. This redistribution dynamically reconfigures the potential field to fill the defect. This capability is considered to be critically important for enhancing the fault tolerance of swarm robotic systems.

### 4.2.3 Parameter Space and Morphological Diversity

Combinations of the morphological parameter $w$ and growth threshold $G$ yielded a wide variety of morphologies, ranging from simple circular clusters to polydactylous extensions and complex network structures. The fact that the parameter range capable of sustaining meaningful configurations is limited is a commonality shared by other artificial life models such as Lenia [17] and Swarm Chemistry [18]. However, a distinctive characteristic of the proposed model is that the "token propagation range $Z$," a variable directly associated with physical communication costs, explicitly dictates the scale of morphology. Future research should explore more advanced internal structuralization, such as organ-like differentiation, through the implementation of multi-layered hierarchical models [16].

### 4.2.4 Challenges toward Engineering Implementation

The engineering implementation of the proposed model requires modules capable of exchanging numerical values with neighbors and storing those values in memory. Regarding the memory overhead, if each cell maintains the accumulated token volume $Sv$ for each aging label $v$ (from 1 to $Z$), each cell must allocate an array of length $Z$. The value $Z = 20$ utilized in this study is well within the implementation capacity of contemporary, low-cost microcontrollers (e.g., those with only a few KB of RAM), suggesting that the model is highly applicable to nano- and micro-robotic systems.

Furthermore, while the current model is evaluated through hexagonal lattice simulations, future work must verify its robustness against "noise"—such as intermittent physical contact or communication latency—when applied to actual micro-robotic swarms. As the token model is based on fundamental arithmetic operations, it is inherently compatible with digital circuits. However, the influence of the aging process precision (elapsed time management)—particularly in the inheritance model—on morphological stability requires further detailed investigation.

## 5. Conclusion

This study has developed a generalized model that enables complex morphogenesis driven solely by adjacent interactions between internal modules, eliminating the need for external spatial computation.

**Research Outcomes**: The challenge of "dependency on external references" inherent in traditional cellular automata models was addressed through the implementation of an "internally restricted equal-distribution token propagation algorithm." Consequently, the model successfully reproduced dynamic morphogenetic processes, including growth, degeneration, and self-division, using only local rules.

**Scholarly Contributions**: This research demonstrated that the token model, which replaces the complex mathematics of reaction-diffusion systems with simple numerical exchanges, can induce Turing instability even within closed domains, thereby functioning effectively as a potential field for morphogenesis.

**Future Outlook**: Future work will seek to trigger the emergence of individuals with more advanced functional segmentation by applying multi-layered or hierarchical architectures to the present model. Furthermore, implementation tests on physical swarm robots (targeting at nano- and micro-scale machines) are necessary to verify self-organization and self-regeneration capabilities under realistic physical constraints.

This framework is expected to contribute to the advancement of autonomous modular robotics without central control and to the field of morphogenetic engineering based on artificial life approaches.

**DECLARATIONS**

**Ethical approval:** Not applicable.

**Consent to participate:** Not applicable.

**Consent to publish:** The author consents to the publication of this work.

**ACKNOWLEDGMENTS**

**Author contributions:** T.I. conceptualized the methodology, software development, formal analysis, and investigation of the study. T.I. also drafted and revised the manuscript, gave final approval for publication, and agreed to be accountable for all aspects of the work.

**Funding:** This research was supported by grants from the Japan Society for the Promotion of Science, KAKENHI Grant Number 23K04283.

**Competing interests:** The author declares no conflict of interest regarding the publication of this paper.

DATA AVAILABILITY

All data supporting the findings of this study are available within the main text and its supplementary materials. Three supplementary videos (reference growth, self-replication, cutting, and regeneration) are presented.

Video 1: **Morphogenesis results under the reference parameter set（Fig.2）**
https://youtu.be/PBNg8WU0tvE

Video 2: **Morphogenesis results of the self-replicating pattern (Fig. 4)**
https://youtu.be/mJ0BRtmIkrg

Video 3: **Morphogenesis results of morphological regeneration patterns (limb extension, amputation, and regeneration) (Fig. 5)**
https://youtu.be/pfUHhLAbJYU

Video 4: **Morphogenesis results of morphological regeneration patterns (mesh expansion, amputation, and regeneration) (Fig. 5)**
https://youtu.be/x2Py1BOFZLc

The source code (HTML + JavaScript) used for the simulations is available on GitHub. The results presented in this paper can be reproduced by changing the UI parameters of the program.

https://github.com/Takeshi-Ishida/Internalized-Morphogenesis-Self-Organizing-Model.git

# Supplementary Materials

**Appendix 1: Basic Morphogenic Behavior**

The time-series transitions of the evaluation indices under the reference parameters are shown in Figure A1.

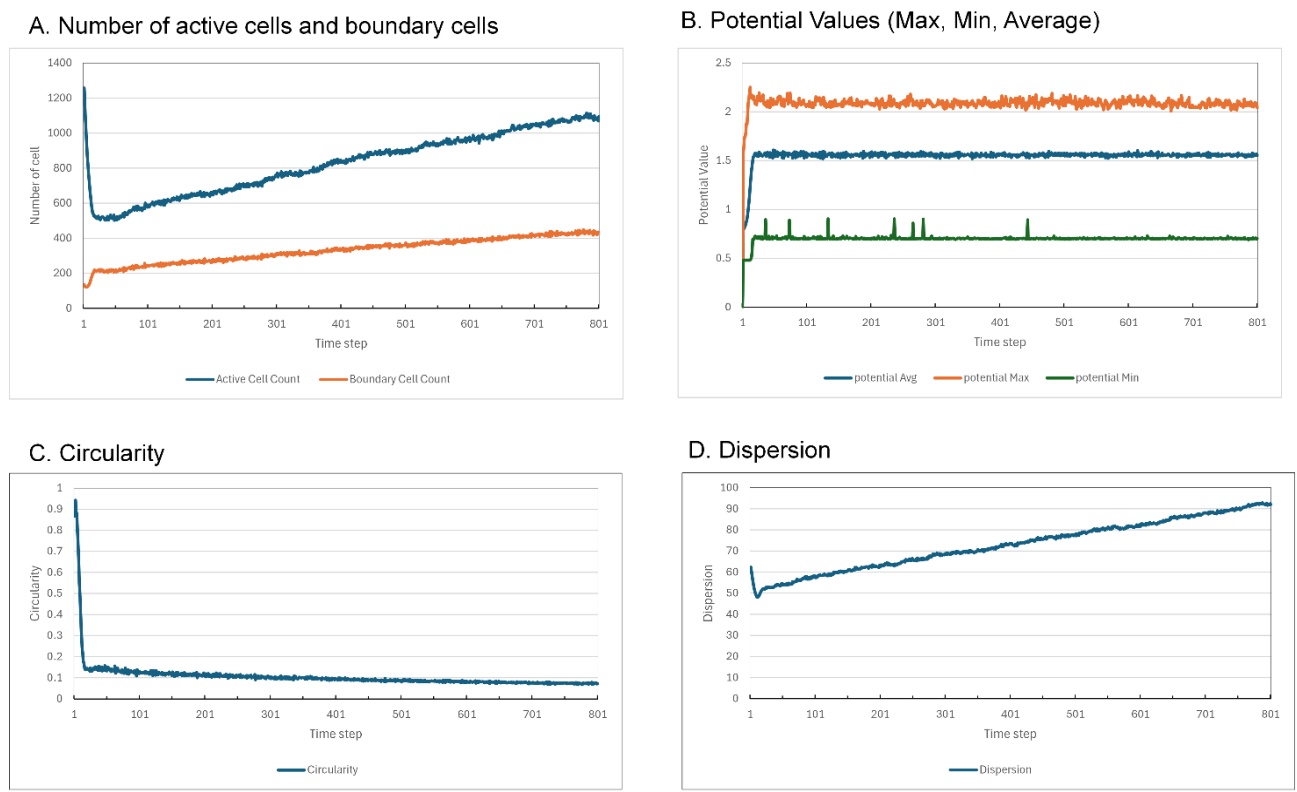

Figure A1: Time-series transitions of evaluation indices under the reference parameters.

**Appendix 2: Parameter Sensitivity and Phase Diagrams; Influence of the *w* Parameter**

Figure A2 shows the results of varying the *w* value relative to the reference parameter set (results at 400 steps). Additionally, the time-series results for *w* = 0.45 are presented in Figure A3. In this case, the boundaries expand, and black spots that are incidentally formed when the morphology collided with the periodic boundary replicate internally and proliferate throughout the domain.

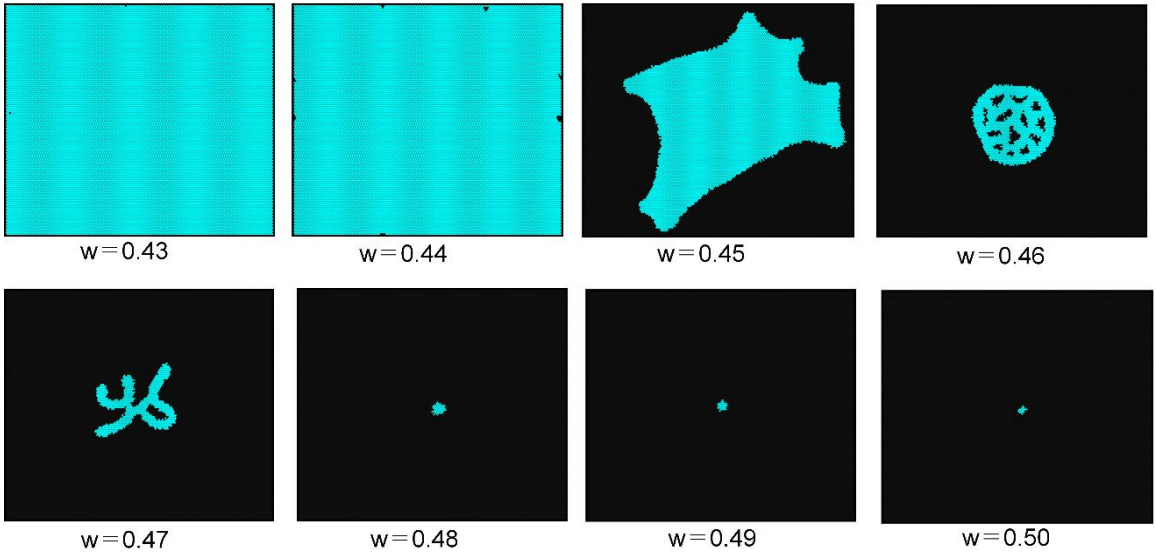

Figure A2: Results of varying the $w$ value relative to the reference parameter set (at 400 steps).

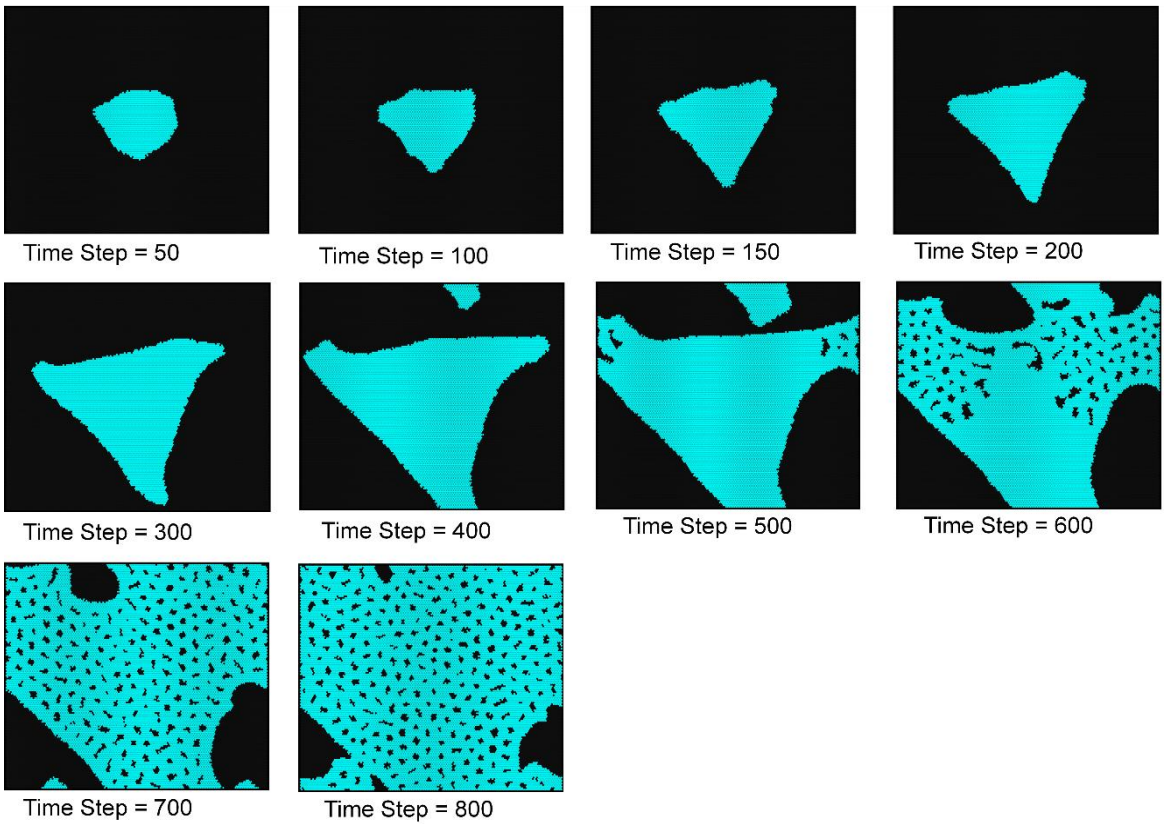

Figure A3: Time-series results for $w = 0.45$; the boundary expands, and black spots formed incidentally upon collision with the periodic boundary replicate and spread internally.

## Appendix 3: Parameter Sensitivity; Influence of Token Propagation Steps (*Z*)

Reducing $Z$ (to $Z \leq 17$) necessitates a corresponding reduction in the setting of $Y$. Simulations revealed that varying $Z$ from 18 to 28 resulted in minimal deviations from the reference case, consistently producing limb-extension patterns.

## Appendix 4: Parameter Sensitivity and Phase Diagrams; Influence of Ranges *X* and *Y*

Figure A4 shows the results of varying the *X* and *Y* values relative to the reference parameter set (results at 400 steps). The reference parameters, $X = 8$ and $Y = 16$, were located at the center of the phase diagram. Keeping $Y = 16$ constant while varying *X*, it was observed that for $X \leq 7$, the initial region contracted and eventually disappeared. Conversely, for $X \geq 9$, the initial region expanded to fill the entire space. Keeping $X = 8$ constant while varying *Y*, for $Y \leq 14$, the initial region expanded, at $Y = 15$, the initial configuration is largely maintained. At $Y = 17$, the initial region contracted with the emergence of small limbs, and at $Y = 18$, it contracted and persisted as a small cluster. In the case of $X = 9$ and $Y = 18$, the initial region expanded incrementally, drawing State 0 voids into the interior, leading to a pattern where State 0 spots underwent replication.

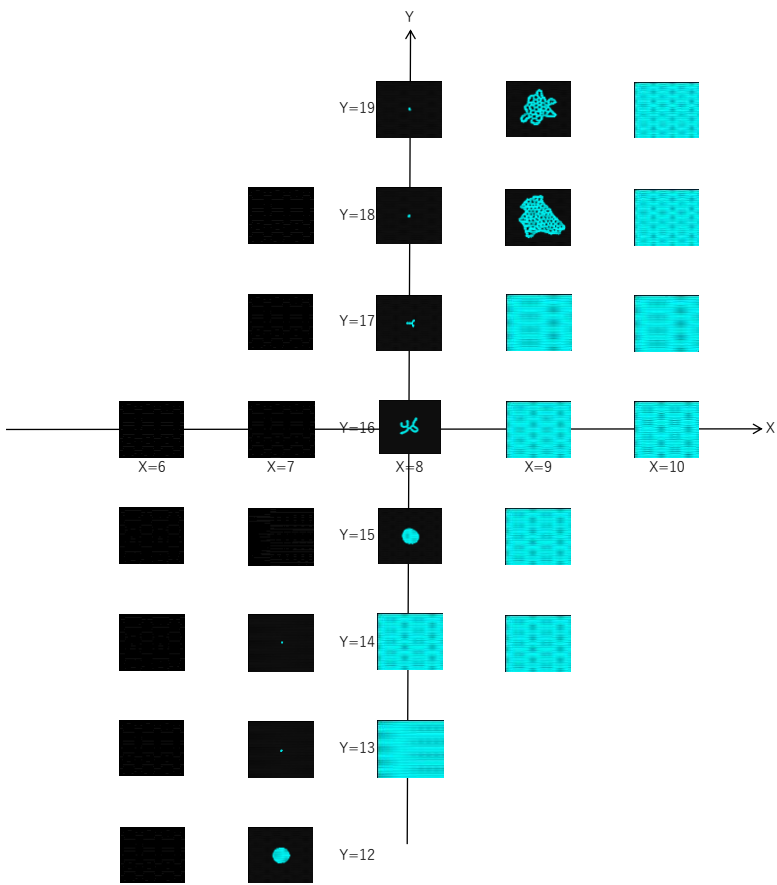

Figure A4: Results of varying *X* and *Y* values relative to the reference parameter set (at 400 steps).

## Appendix 5: Configuration of Module Growth on Boundaries; Differences in Results between "Boundary Only" and "Including Internal Domain"

Figure A5 presents the time-series results obtained when the module growth configuration for boundary processing is set to "Including Internal Domain" under the reference parameter set. The morphology expanded while maintaining a circular shape, leading to the formation of internal black spots. These internal spots changed dynamically, resulting in an unstable pattern.

Figure A6 shows the results of varying the $w$ value (at 400 steps) for the reference parameter set combined with the "Including Internal Domain" configuration. For $w$ in the range of 0.43 to 0.46, the region expands, and the interior exhibits a flickering state of black spots. As $w$ increases, the rate of regional expansion decreases. At $w = 0.480$, the region contracts rapidly, leaving a small cluster. At $w = 0.490$, this region immediately disappeared.

Figure A7 presents the results of varying the growth threshold $G$ (at 400 steps) for the reference parameter set with the "Including Internal Domain" configuration. Figure A8 shows the results of varying the survival threshold $R$ (at 400 steps) under the same conditions.

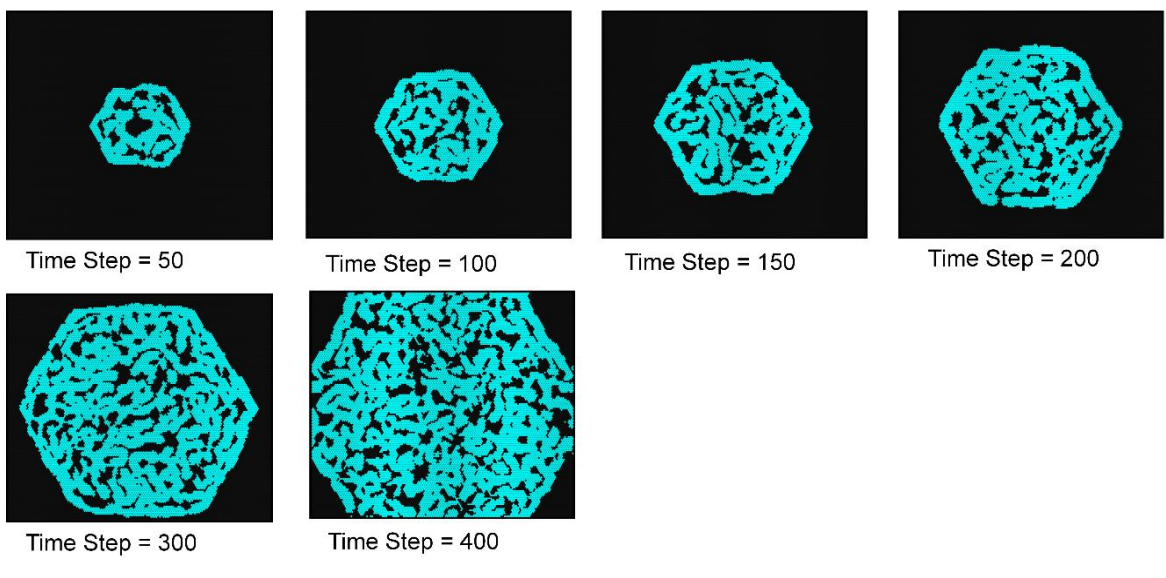

Figure A5: Time-series results for the reference parameter set with the "Including Internal Domain" configuration.

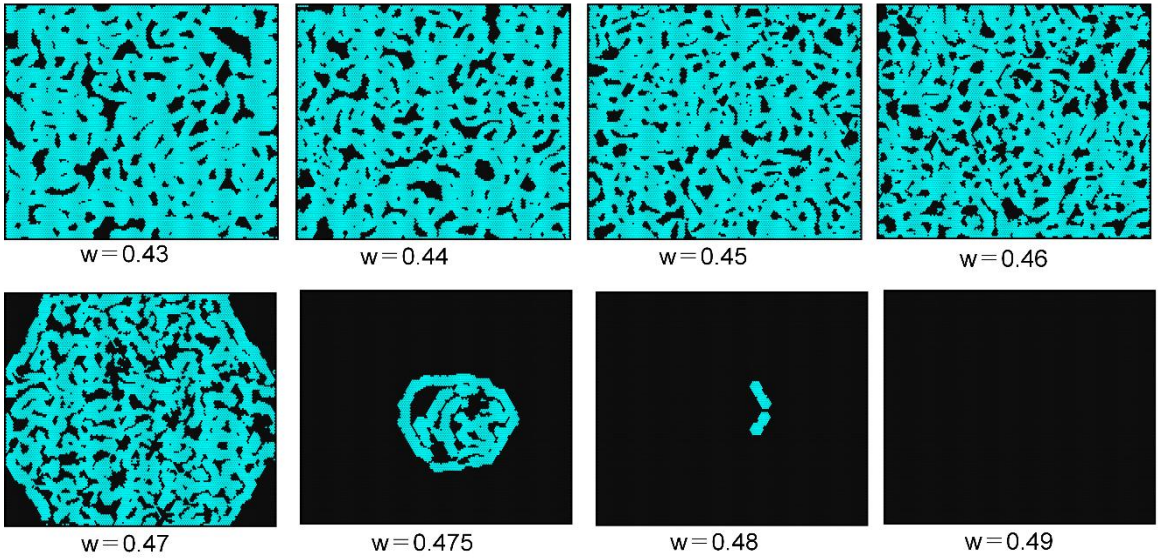

Figure A6: Results of varying the *w* value for the reference parameter set with the "Including Internal Domain" configuration (at 400 steps).

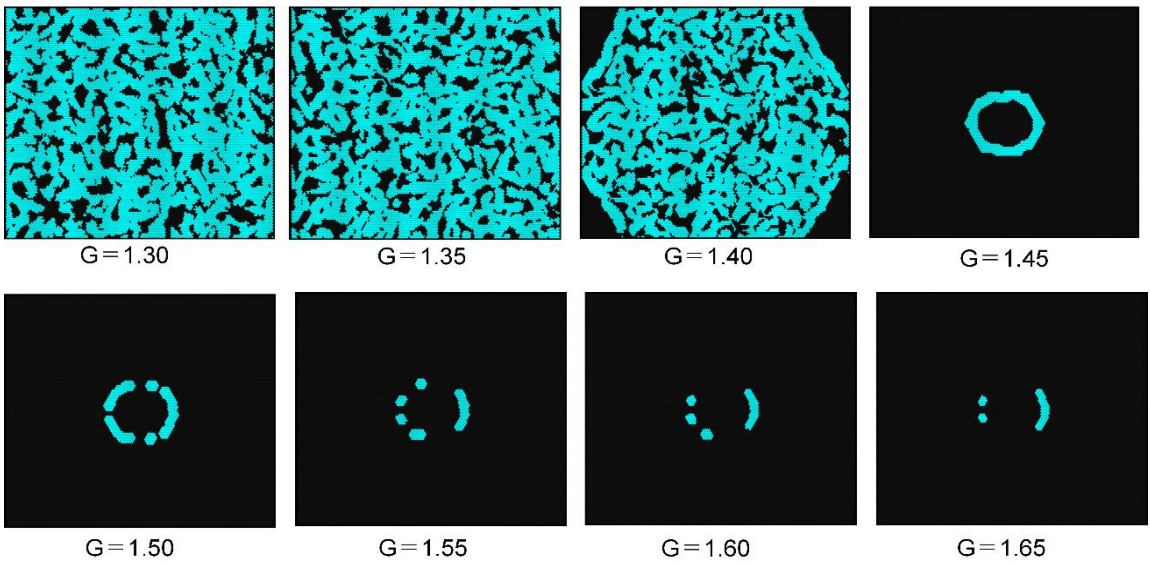

Figure A7: Results of varying the growth threshold *G* for the reference parameter set with the "Including Internal Domain" configuration (at 400 steps).

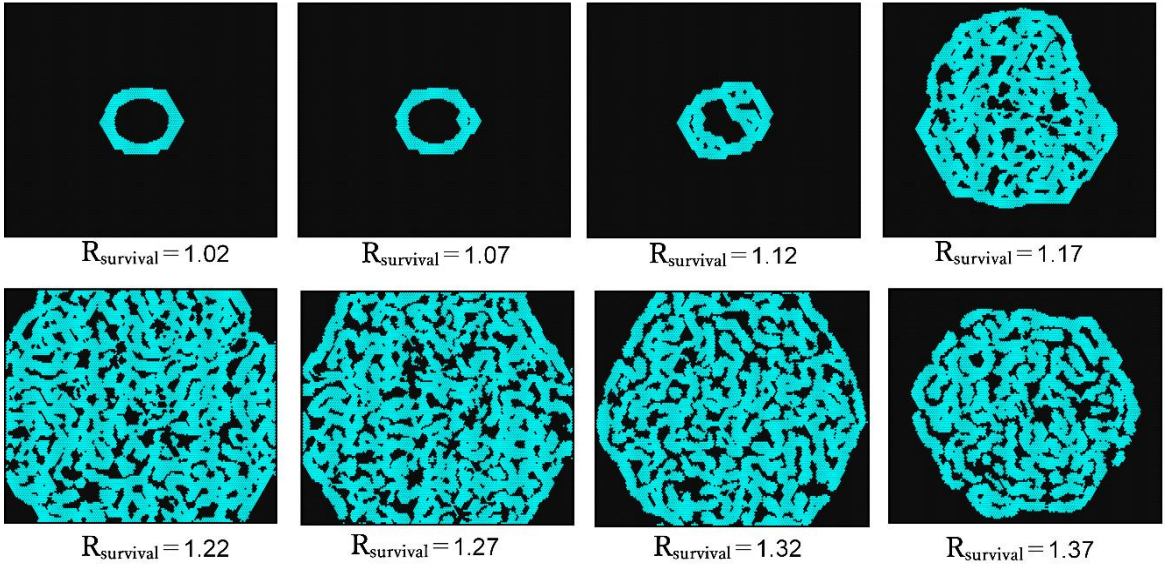

Figure A8: Results of varying the survival threshold *R* for the reference parameter set with the "Including Internal Domain" configuration (at 400 steps).

**Appendix 6: Summary of Results for the "Inheritance Model"**

The following section presents the results obtained using the "inheritance model," which allows tokens to persist from previous steps. Using the same parameter set as in the reference case, the transfer rate *k* was varied.

Figure A9 illustrates the results of varying transfer rate *k* relative to the reference parameter set in the inheritance model (results at 400 steps). When the transfer rate was 0, the results were identical to those of the basic model, which is characterized by limb extension. At $k = 0.02$, the limbs continued to extend, although the growth rate was slightly reduced. For $k = 0.04$, 0.06, and 0.07, the initial region contracts into a small cluster, whereas the limbs continue to extend. At $k = 0.08$, the morphology disappears, and all transfer rates higher than this result in extinction patterns.

Figure A10 shows the results for the inheritance model when $G = 1.30$ and $R = 1.17$ are set, depicting the effects of varying the transfer rate *k* (results at 400 steps). At transfer rates of $k = 0.06$ and 0.08, mesh-like growth patterns emerge, although the boundaries exhibit slight instability.

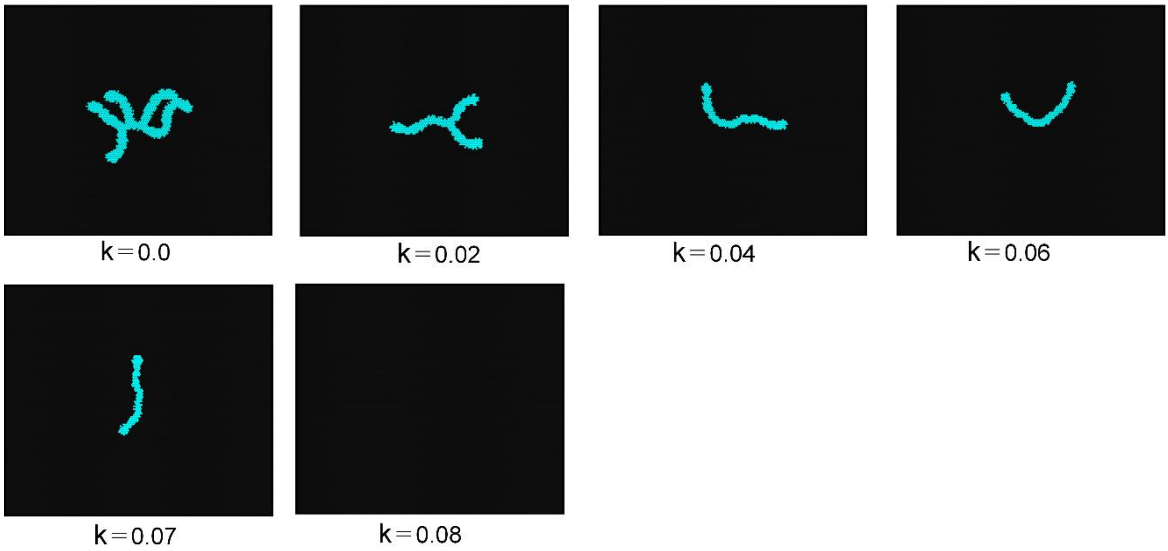

Figure A9: Results of varying the transfer rate *k* for the inheritance model under the reference parameter set (at 400 steps).

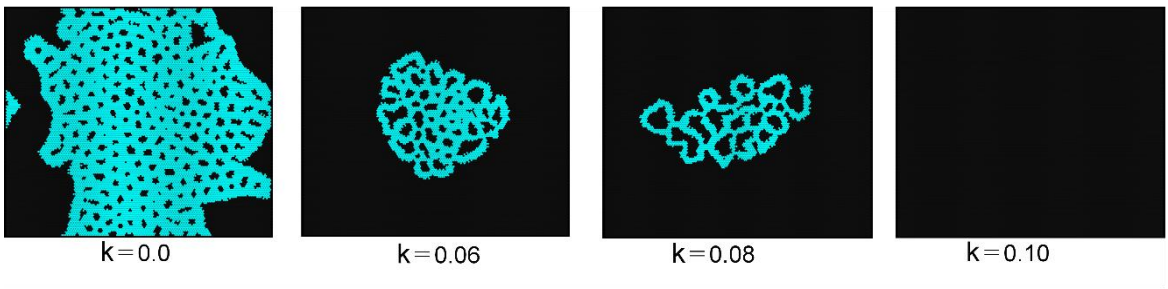

Figure A10: Results of varying the transfer rate *k* for the inheritance model with $G = 1.30$ and $R = 1.17$ (at 400 steps).

**Appendix 7: Results of Self-Replication Patterns; Self-Replication Pattern in the "Inheritance Model" (Transfer Rate *k* = 0.02)**

A parameter search revealed that self-replication patterns emerged under the following settings relative to the reference parameters: $w = 0.50$, $X = 8$, $G = 1.37$, $R = 1.35$, $Ro = 1.77$.

When $Ro = 1.77$, the morphology typically formed a mesh-like structure, but fragmentation occasionally occurred, leading to a self-replication pattern (Figure A11). At $Ro = 1.76$, extinction patterns were common, although self-replication occasionally occurred. At $Ro = 1.75$, the result was an extinction pattern. In the case of $Ro = 1.78$, a stable mesh pattern forms. A change of only 0.01 in $Ro$ leads to a significant shift in the resulting pattern. Further detailed investigation of these settings is required in future studies.

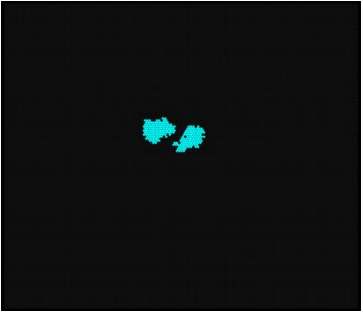

Figure A11: Self-replication pattern in the "inheritance model" (transfer rate $k = 0.02$).

**Appendix 8: Time-Series Transitions of Indices for Self-Regeneration Patterns**

Figure A12 illustrates the time-series results of the evaluation indices for a self-regeneration pattern involving limb extension, amputation at 200 steps, and subsequent recovery (parameters: $G = 1.42$, $R = 1.2$; all other parameters are identical to the reference set).

Figure A13 presents the time-series results of the evaluation indices for a self-regeneration pattern involving mesh-region expansion, amputation at 200 steps, and subsequent recovery (parameters: $G = 1.30$, $R = 1.2$; all other parameters are identical to the reference parameter set).

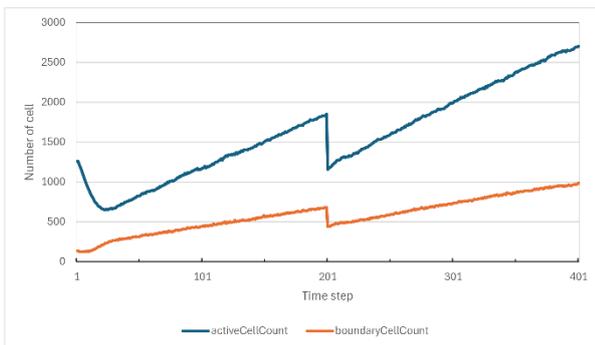
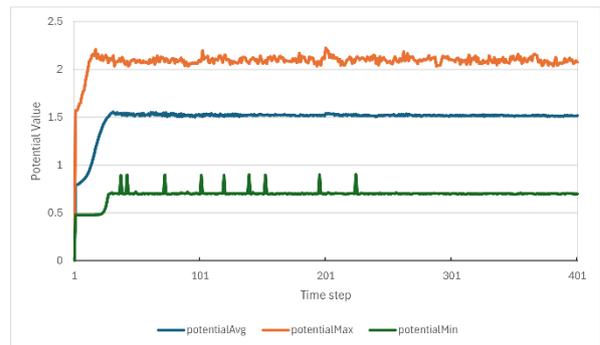
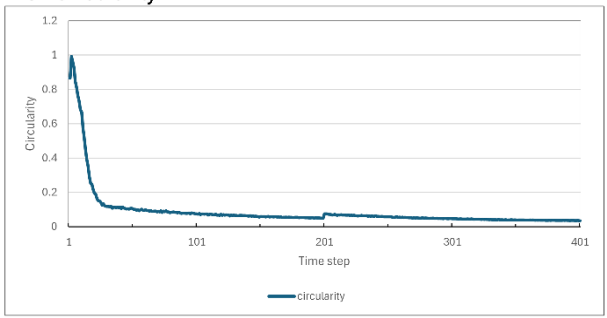
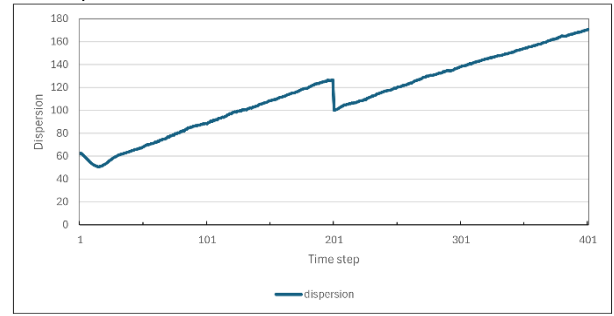

Figure A12: Time-series results of evaluation indices for the self-regeneration pattern (limb extension, amputation at 200 steps, and subsequent regeneration) ($G = 1.42$, $R = 1.2$; other parameters identical to the reference set).

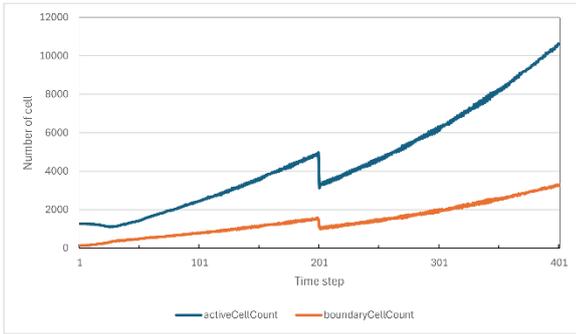
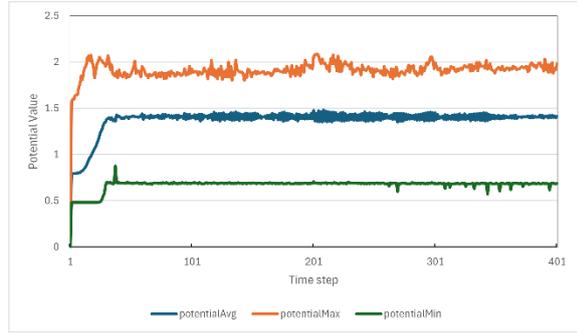
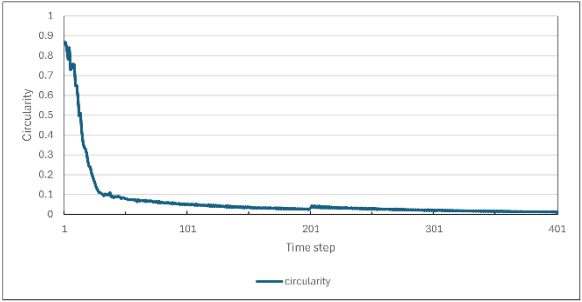
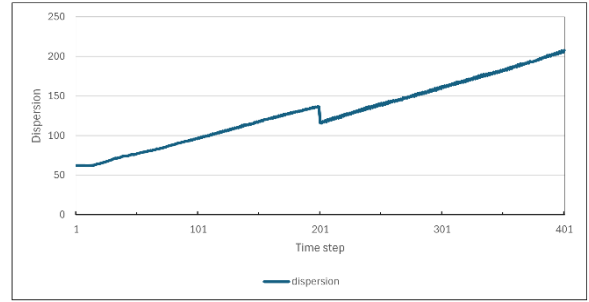

Figure A13: Time-series results of evaluation indices for the self-regeneration pattern (mesh-region expansion, amputation at 200 steps, and subsequent regeneration) ($G = 1.30$, $R = 1.2$; other parameters identical to the reference parameter set).

### Appendix 9: Pseudocode of the Basic Model

**Definitions:**

$Grid$: The hexagonal lattice space.

$S_i \in \{0, 1\}$: State of cell $i$ (0: Empty, 1: Active Module).

$N(i)$: Set of 6 neighboring cells of cell $i$.

$T_{i, v}$: Amount of tokens with label (age) $v$ accumulated in cell $i$.

$Z$: Maximum token propagation steps.

$X, Y$: Parameters for potential calculation ranges ($X < Y$).

$w$: Weight parameter for inhibition.

$R_{surv}$, $R_{over}$: Thresholds for survival (lower limit) and overcrowding (upper limit).

$G$: Threshold for growth.

**Algorithm 1: Main Simulation Loop**

Initialize Grid with initial state (e.g., a circle of Active cells)

```
t <- 0
While Simulation is Running do
    // Step 1: Token Propagation and Potential Calculation
    CalculatePotentials(Grid)

    // Step 2: Update Morphology (Death and Growth)
    UpdateMorphology(Grid)

    t <- t + 1
End While
```

**Algorithm 2: Token Propagation and Potential Calculation**

```
Procedure CalculatePotentials(Grid)
    // 1. Reset tokens for the new step
    For each cell i in Grid do
        Initialize T_{i, v} = 0 for all v in {1, ..., Z}
        If S_i = 1 then
            // Generate initial token (age v=1)
            T_{i, 1} <- 1.0
        End If
    End For

    // 2. Propagate tokens up to Z steps
    // Let TempT_{i} be the tokens currently residing in cell i to be distributed
    For v = 1 to Z - 1 do
        For each cell i with S_i = 1 do
            If T_{i, v} > 0 then
                AmountToSend <- T_{i, v} / 6
                For each neighbor n in N(i) do
                    If S_n = 1 then
                        // Token is received and ages by 1
                        T_{n, v+1} <- T_{n, v+1} + AmountToSend
                    Else
                        // Token dissipates into empty space (Boundary Sink)
                        Discard AmountToSend
                    End If
                End For
            End If
```

            End For

        End For

        // 3. Calculate Potential $P_i$ for each cell

        For each cell i with $S_i = 1$ do

            SumX <- Sum($T_{\{i, v\}}$) for v = 1 to X

            SumY <- Sum($T_{\{i, v\}}$) for v = 1 to Y

            $P_i$ <- SumX - (SumY * w)

        End For

End Procedure

### Algorithm 3: Morphological Update (Death and Growth)

Procedure UpdateMorphology(Grid)

    NextGrid <- Copy(Grid) // Synchronous update buffer (optional, depending on implementation specifics)

    // 1. Shrinkage (Death) Rule

    // Applies to ALL active cells

    For each cell i with $S_i = 1$ do

        If $P_i < R_{\{surv\}}$ OR $P_i > R_{\{over\}}$ then

            NextGrid[i] <- 0 // Cell dies

        End If

    End For

    // Update Grid to reflect deaths before calculating growth

    Grid <- NextGrid

    // 2. Growth Rule

    // Applies to Boundary cells only

    Candidates <- Empty list

    For each cell i with $S_i = 1$ do

        If IsBoundary(i) AND $P_i >= G$ then

            // Identify empty neighbors

            EmptyNeighbors <- {n in N(i) | $S_n = 0$}

            If EmptyNeighbors is not empty then

                Target <- SelectRandom(EmptyNeighbors)

                Add (Target, i) to Candidates

            End If

        End If

```
        End For

        // Resolve Conflicts and Execute Growth
        // If multiple cells try to grow into the same Target, select one randomly
        For each unique Target t in Candidates do
                Grid[t] <- 1 // New module is created
        End For
End Procedure
```